\pdfoutput=1
\documentclass[11pt]{article}
\usepackage[hyperref]{emnlp2020}
\usepackage{times}
\usepackage{latexsym}
\usepackage{booktabs}
\usepackage{amsthm,amsmath,amssymb}
\usepackage{graphicx}
\usepackage{makecell}
\usepackage{xcolor}
\usepackage{soul}
\usepackage{ulem}
\usepackage{todonotes}
\usepackage{bm}
\usepackage{color}
\usepackage{multirow}

\usepackage{mathrsfs}
\usepackage{microtype}

\newcommand{\set}{\mathcal}

\makeatletter
  \newcommand\reduline{\bgroup\markoverwith{\textcolor{red}{\rule[-0.5ex]{2pt}{0.4pt}}}\ULon}
  \UL@protected\def\blueuwave{\leavevmode \bgroup 
    \ifdim \ULdepth=\maxdimen \ULdepth 3.5\p@
    \else \advance\ULdepth2\p@ 
    \fi \markoverwith{\lower\ULdepth\hbox{\textcolor{blue}{\sixly \char58}}}\ULon}
  \UL@protected\def\yellowdotuline{\leavevmode \bgroup 
    \UL@setULdepth
    \ifx\UL@on\UL@onin \advance\ULdepth2\p@\fi
    \markoverwith{\begingroup
      \lower\ULdepth\hbox{\kern.06em \textcolor{yellow}{.}\kern.04em}%
      \endgroup}%
    \ULon}
  \UL@protected\def\greendashuline{\leavevmode \bgroup 
    \UL@setULdepth
    \ifx\UL@on\UL@onin \advance\ULdepth2\p@\fi
    \markoverwith{\kern.13em
    \vtop{\color{green}\kern\ULdepth \hrule width .3em}%
    \kern.13em}\ULon}
\makeatother

\usepackage{microtype}

\title{Dynamic Schema Graph Fusion Network for \\ Multi-Domain Dialogue State Tracking}

\author{Yue Feng$^\dagger$ \quad Aldo Lipani$^\dagger$ \quad Fanghua Ye$^\dagger$ \quad Qiang Zhang$^\ddagger$ \thanks{Work in part done while at University College London.} \quad Emine Yilmaz$^\dagger$\\
  $^\dagger$University College London, London, UK\\
  $^\ddagger$Zhejiang University, Hangzhou, China\\
  $^\dagger$ {\texttt{\{yue.feng.20,aldo.lipani,fanghua.ye.19,emine.yilmaz\}@ucl.ac.uk}} \\
  $^\ddagger$\texttt{qiang.zhang.cs@zju.edu.cn}\\}

\begin{document}
\maketitle
\begin{abstract}
Dialogue State Tracking (DST) aims to keep track of users' intentions during the course of a conversation. In DST, modelling the relations among domains and slots is still an under-studied problem. Existing approaches that have considered such relations generally fall short in: (1) fusing prior slot-domain membership relations and dialogue-aware dynamic slot relations explicitly,
% (2) explaining dialogue-aware dynamic slot relations explicitly; 
and (2) generalizing to unseen domains. To address these issues, we propose a novel \textbf{D}ynamic \textbf{S}chema \textbf{G}raph \textbf{F}usion \textbf{Net}work (\textbf{DSGFNet}), 
which generates a dynamic schema graph to explicitly fuse the prior slot-domain membership relations and dialogue-aware dynamic slot relations. It also uses the schemata to facilitate knowledge transfer to new domains.
% which is able to generate dynamic schema graph to explicitly fuses static slot relations and dynamic slot relations, and utilizes schema to facilitate knowledge transfer to new domains.
% which explicitly fuses static slot relations and dynamic slot relations using dynamic schema graph, and utilizes schema to facilitate knowledge transfer to new domains. %with the use of a unified schema-agnostic model. 
 DSGFNet consists of a dialogue utterance encoder, a schema graph encoder, a dialogue-aware schema graph evolving network, and a schema graph enhanced dialogue state decoder. Empirical results on benchmark datasets (i.e., SGD, MultiWOZ2.1, and MultiWOZ2.2), show that DSGFNet outperforms existing methods.
 
 %including unseen domains on SGD, all domains on SGD, MultiWOZ2.1, and MultiWOZ2.2, show that DSGFNet outperforms the existing methods.
%  by 6.6\%, 2.6\%, and 2.5\%, respectively. The results verify the effectiveness of the dynamic schema graph in DST.
% We first evaluate DSGFNet on the most challenging benchmark SGD. Experimental results show that DSGFNet achieves new state-of-the-art performance measured by JGA, and outperforming the runner-up by 2.0 points (32.1\% vs 30.1\%). Then, we further evaluate on much smaller benchmarks MultiWOZ2.1 and MultiWOZ2.2, the proposed method achieves consistent improvements.~\QZ{Why not try Fanghua's cleaner MultiWoZ2.4 dataset?}
\end{abstract}

\begin{figure}[!t]
\centering
\includegraphics[width=0.49\textwidth]{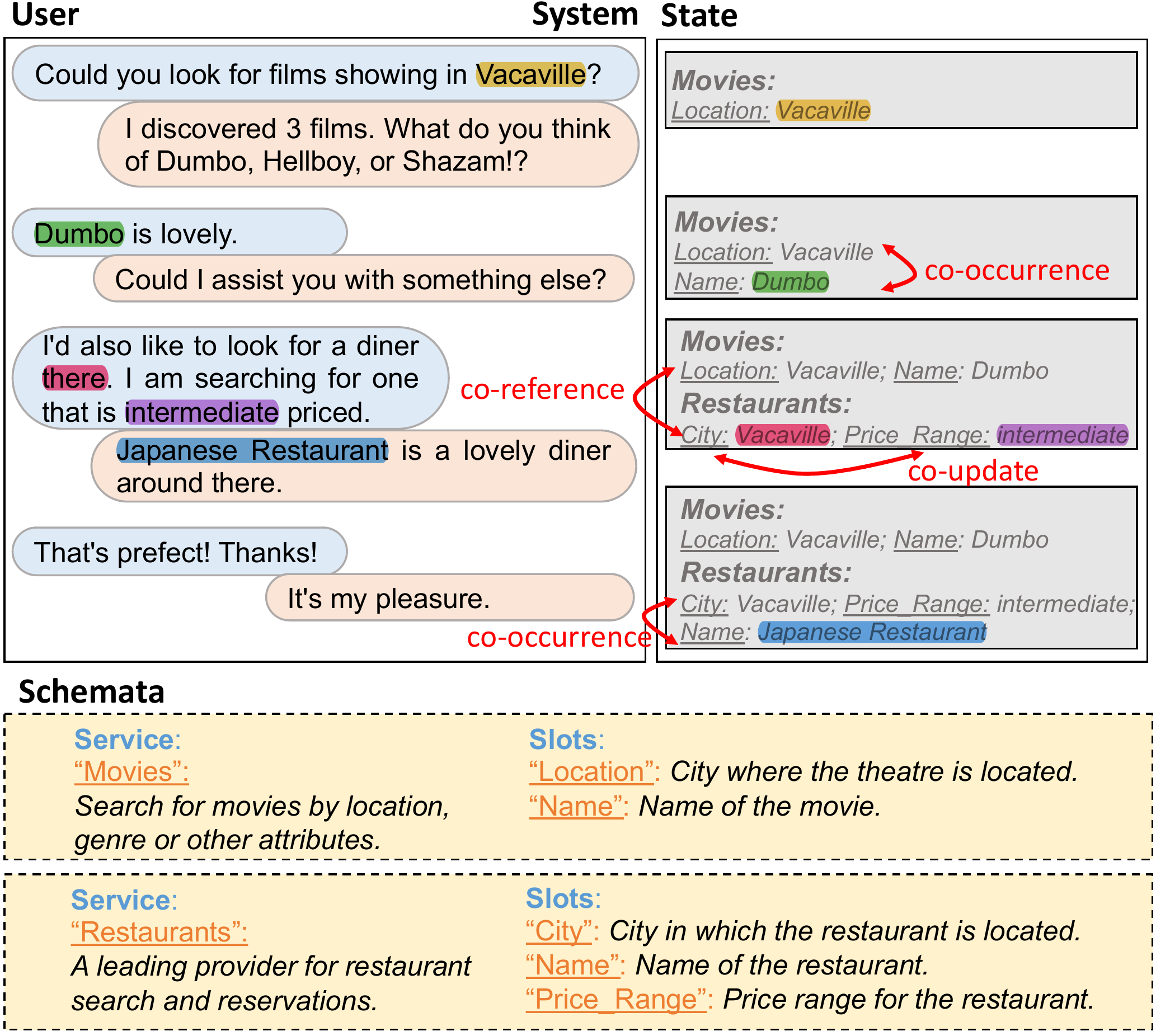}
\caption{An example of DST. Given the schemata for all domains, the slot values are extracted from the user and system utterances (e.g., spans highlighted with the same color in the figure). The dialogue state of each turn is represented as a set of slot-value pairs. Among the domains and slots, there are prior slot-domain membership relations which are expressed in the predefined schemata, and also dialogue-aware dynamic slot relations which depend on the dialogue context (e.g., co-reference, co-update, and co-occurrence).
}
% \vspace{-10pt}
\label{fig:example}
\end{figure}

\section{Introduction}

Task-oriented dialogue systems can help users accomplish different tasks~\citep{huang2020challenges}, such as flight reservation, food ordering, and appointment scheduling. Conventionally, task-oriented dialogue systems consist of four modules~\citep{zhang2020recent}: natural language understanding (NLU), dialogue state tracking (DST), dialogue manager (DM), and natural language generation (NLG). In this paper, we will focus on the DST module.
The goal of DST is to extract users' goals or intentions as dialogue states and keep these states updated over the whole dialogue. 
In order to track users' goals, we need to have a predefined domain knowledge referred to as a schema, which consists of slot names and their descriptions.
% The goal of DST is to extract user goals/intentions from their inputs and update this information over the continuation of the dialogue.
Figure~\ref{fig:example} gives an example of DST in a sample dialogue.

Many models have been developed for DST due to its importance in task-oriented dialogue systems. 
% State-of-the-art DST models can be grouped into two categories, depending on whether or not they predict slot values independently of their relations. 
% {\color{red} The categorization is not standard}
% The independent prediction of slot values requires a predefined schema, where all slots and part of their candidate values are known. Such a schema is defined for each domain as an ontology containing slot names and their descriptions.
% The independent prediction of slot values simplifies the DST problem into a generation problem for each slot, by either directly producing a slot from a given schema or extracting a value from the dialogue context
Traditional approaches use deep neural networks or pre-trained language models to encode the dialogue context and infer slot values from it~\citep{zhong2018global,ramadan2018large,wu2019transferable,ren2019scalable,zhang2019find,hu2020sas,gao2020machine,zhang2019find,zhang2020probabilistic}.
% ~\cite{mrkvsic2017neural, nouri2018toward, lee2019sumbt,lei2018sequicity, gao2019dialog}.
These models predict slot values without considering the relations among domains and slots. However, domains and slots in a dialogue are unlikely to be entirely independent, and ignoring the relations among domains and slots may lead to sub-optimal performance.
To address this issue, several recent works have been proposed to model the relations among domains and slots in DST. 
Some of them introduce predefined schema graphs to incorporate prior slot-domain membership relations, which are defined based on human experience in advance~\citep{chen2020schema, zhu2020efficient}. 
The others use an attention mechanism to capture dialogue-aware dynamic slot relations~\citep{feng2020sequence,heck2020trippy}. The dialogue-aware dynamic relations are the logical relations of slots across domains, which are highly related to specific dialogue contexts. 
% {\color{red}I would suggest that you remove categorization of the methods but say that most existing methods fail to consider slot relations. Although some models consider the relation, they still have severe limitations}

% As slots in a practical dialogue are unlikely to be entirely independent, involving slot relations can significantly enhance the performance of DST. 
% However, existing models involving relations among domains and slots in DST suffer from three major issues: %
% (1) They fail to fuse prior static relations and dialogue-aware dynamic relations. Prior static relations are defined by human experience in advance, such as slot-domain membership relations.
However, existing DST models that involve the relations among domains and slots suffer from two major issues:
(1) They fail to fuse the prior slot-domain membership relations and dialogue-aware dynamic slot relations explicitly; and 
% The prior slot-domain membership relations are defined based on human experience in advance. The dialogue-aware dynamic relations are the logical relations of slots across domains, which are highly related to specific dialogue contexts. 
% Static slot relations are based on prior knowledge about slot relevance, which are defined by human experience in advance. 
% While dynamic slot relations are constantly occurring in a specific dialogue context. %It is essential to analyze dynamic logic relations based on prior knowledge. 
% Dialogue-aware dynamic relations are logical relation of slots across domains in dialogue context, which are highly related to specific dialogue context. 
% (2) They fail to explain the dialogue-aware dynamic slot relations explicitly. Ignoring the relation meanings may lead to inappropriate utilization and integration of the relations.
(2) They fail to consider their generalizability to new domains. In practical scenarios, task-oriented dialogue systems need to support a large and constantly increasing number of new domains. 
% {\color{red}unseen schema? doesn't your approach require the schemata of new domains are available }
% Since slot relations are implicitly represented in the trained models, it is hard to interpret and present the obtained slot relations.
%Previous works fail in capturing slot relations in new domains.
% \QZ{Not clear: are you talking about generalization to new domains with unseen schema or scalability when there is a large amount of data? }
% \YF{Can I use scalability to express generalization to new domains with unseen schema? \color{violet} Aldo: I think you mean generalizability, scalability may be confusing because it can be interpreted in different ways.}
% {\color{red}for the second point, what's the meaning of it? why do we care about the explanation of slot relations? besides, some methods modelling static slot relations have strong interpretability.}

To tackle these issues, we propose a novel approach named DSGFNet (Dynamic Schema Graph Fusion Network). 
%For the first issue, DSGFNet dynamically updates the predefined slot-domain membership relation schema graph with the dialogue-aware dynamic slot relations. 
%
For the first issue, DSGFNet dynamically updates the schema graph consisting of the predefined slot-domain membership relations with the dialogue-aware dynamic slot relations. 
To incorporate the dialogue-aware dynamic slot relations explicitly, DSGFNet adds three new edge types to the schema graph: %that originally contains %just the slot-domain membership relations, including 
\textit{co-reference relations}, \textit{co-update relations}, and \textit{co-occurrence relations}. 
% Co-reference relations occur when a slot value has been mentioned earlier in the dialogue and has been assigned to another slot. Co-update relations occur when slot values are updated together at the same dialogue turn. Co-occurrence relations occur when slots with a high co-occurrence probability in a large dialogue corpus appear together in the current dialogue. Examples of these slot relations are shown in Figure~\ref{fig:example}. 
For the second issue, to improve its generalizability, DSGFNet employs a unified model containing schema-agnostic parameters to make predictions.
%
% \add{It seems this paragraph indicates the second contribution, but it is not clear how the second point in Line172 can be seen as a contribution. You need to explain it here.}
%

Specifically, our proposed DSGFNet comprises of four components:
a \textit{BERT-based dialogue utterance encoder} to contextualize the current turn dialogue context and history, a \textit{BERT-based schema graph encoder} to generalize to unseen domains and model the prior slot-domain membership relations on the schema graph,
a \textit{dialogue-aware schema graph evolving network} to augment the dialogue-aware dynamic slot relations on the schema graph, 
and a \textit{schema graph enhanced dialogue state decoder} to extract value spans from the candidate elements considering the evolved schema graph.
% Experimental results on three benchmark datasets (SGD, MultiWOZ 2.1, MultiWOZ 2.2) show DSGFNet significantly outperforms state-of-the-art baselines, and verify the effectiveness of the dynamic schema graph in DST.

% Aldo: We don't need to say this in the introduction
%Experimental results on benchmark datasets show that DSGFNet performs better than the baselines on 
%SGD, in multi-domain dialogue with unseen schema in the test set, is superior to the baselines on MultiWOZ2.2, and MultiWOZ2.1, in multi-domain dialogue without unseen schema. %
%An ablation analysis demonstrates the effect of each component in DSGFNet and extensive discussions show the usefulness of the slot relations in DST. %the effects of slot relations on DST~\QZ{ans shows ...}.

The contributions of this paper can be summarized as follows:
\begin{itemize}
  \item We improve DST by proposing a dynamic, explainable, and general schema graph which explicitly models the relations among domains and slots based on both prior knowledge and the dialogue context, no matter whether the domains and slots are seen or not.
%   \item To fuse static slot relations and dynamic slot relations, 
  \item We develop a fusion network, DSGFNet, which effectively enhances DST generating a schema graph out of the combination of prior slot-domain membership relations and dialogue-aware dynamic slot relations. %, and fuses the dialogue context to enhance DST.
%   \item We effectively fuse the proposed schema graph in DSTs {\color{red} feel like this sentence is not clear} and leverage a BERT-based graph neural network via the attention mechanism to aggregate slot relations. {\color{red} do you mean GAT?}
 \item We conduct extensive experiments on three benchmark datasets %(i.e., unseen domains on SGD, all domains on SGD, MultiWOZ2.1, and MultiWOZ2.2) 
 (i.e., SGD, MultiWOZ2.1, and MultiWOZ2.2) to demonstrate the superiority of DSGFNet\footnote{The code is available at \url{https://github.com/sweetalyssum/DSGFNet}.} and the importance of the relations among domains and slots in DST.
%  \QZ{ Isn't this a contribution?}
\end{itemize}

\begin{figure*}[!th]
\centering
\includegraphics[width=0.91\textwidth]{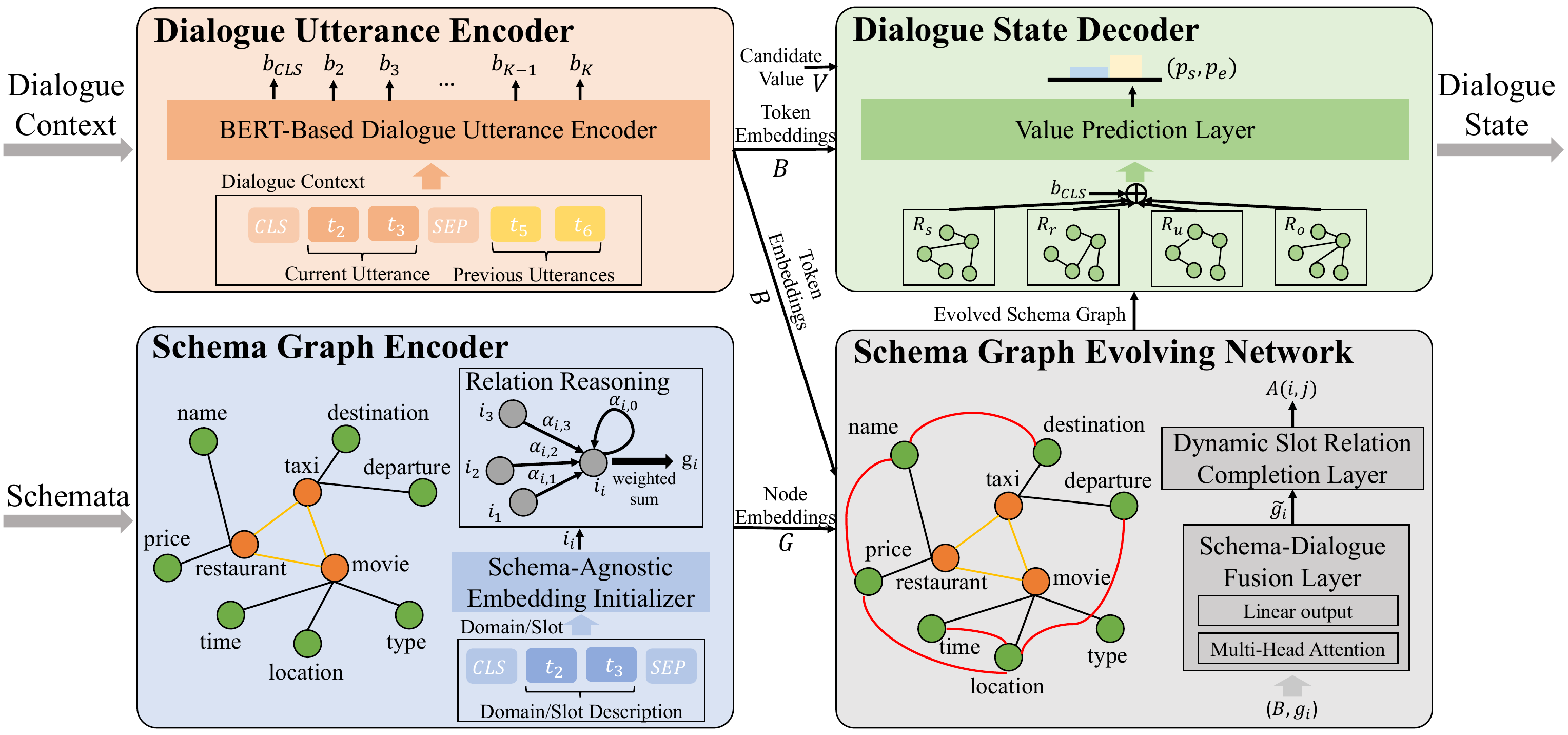}
\caption{The architecture of DSGFNet, which contains a dialogue utterance encoder, a schema graph encoder, a schema graph evolving network, and a dialogue state decoder.}
\vspace{-1em}
\label{fig:framework}
\end{figure*}

\section{Related Work}
%Since DST is a core component in task-oriented dialogue systems, a large amount of research on DST has been proposed. 
% Traditional statistical approaches rely on semantics extracted by the natural language understanding module or complex domain-specific lexicons to predict the dialogue state~\citep{yu2015constrained,wen2017network, rastogi2018multi}. These methods usually suffer from poor generalizability and sub-optimal performance. They are also vulnerable to lexical and morphological variations. Recently, most works about DST focus on encoding dialogue context with deep neural networks (such as CNN, RNN, LSTM-RNN, etc.) in lieu of heuristic features~\citep{zhong2018global,ramadan2018large,ren2019scalable,zhang2019find,hu2020sas,gao2020machine}. Besides, 
Recent DST approaches mainly focus on encoding the dialogue contexts with deep neural networks (e.g., convolutional and recurrent networks) and inferring the values of slots independently~\citep{zhong2018global,ramadan2018large,wu2019transferable,ren2019scalable,zhang2019find,hu2020sas,gao2020machine}. %
With the prevalence of pre-trained language models, such as BERT~\citep{devlin2018bert} and GPT-2~\citep{radford2019language}, a great variety of DST approaches have been developed on top of these pre-trained models
% which predict the value of each slot separately as well
~\citep{zhang2019find,zhang2020probabilistic,lin2020mintl}. The relations among domains and slots are not considered in the above approaches. However, the prior slot-domain membership relations can facilitate the sharing of domain knowledge and the dialogue-aware dynamic slot relations can conduce dialogue history understanding.
% as manifested in the schemata. 
Ignoring these relations may lead to sub-optimal performance.

% all the approaches above ignore considering slots relations that can also manifest across schemata. {\color{red} this sentence is redundant, should be removed, and say slot relations are important}

% With the prevalence of pre-training language models, such as BERT~\citep{devlin2018bert} and GPT-2~\citep{radford2019language}, a great variety of approaches on top of these pre-trained models have been proposed~\citep{chao2019bert,noroozi2020fast,gulyaev2020goal,hosseini2020simple,kim2019efficient,lee2019sumbt,lin2020mintl,shan2020contextual,wang2020fast,wu2020tod,heck2020trippy,ham2020end}.~\add{And what happened then?} All the methods mentioned above cannot model the relations among slots.

To fill in this gap, several new DST approaches, which involve the relations among domains and slots, have been proposed. 
Some of them leverage a graph structure to capture the slot-domain membership relations~\citep{lin2021knowledge,chen2020schema,zhu2020efficient,zeng2020multi,ouyang2020dialogue}. Specifically, a predefined schema graph is employed to represent the slot-domain membership relations. However, they fail to incorporate the dialogue-aware dynamic slot relations into the schema graph. 
The other approaches utilize the attention mechanism to learn dialogue-aware dynamic slot relation features in order to facilitate information flow among slots~\citep{zhou2019multi,feng2020sequence,heck2020trippy, hu2020sas,ye2021slot}. However, these approaches ignore the slot-domain membership relations defined by prior knowledge. Since both the prior slot-domain membership relations and dialogue-aware dynamic slot relations can enhance DST performance, our approach is developed to combine them in an effective way.
% Moreover, to enhance the interpretability, we explicitly represent slot relations in our approach. , which can help propagate information among slots

Given that a deployed dialogue system may encounter an ever-increasing number of new domains that have limited training data available, the DST module should be capable of generalizing to unseen domains. Recent DST approaches have focused on using zero-shot learning to achieve this goal~\citep{rastogi2019towards,noroozi2020fast}. These approaches exploit the natural language descriptions of schemata to transfer knowledge across domains. However, they ignore the relations among domains and slots. In this work, we propose a unified framework to fuse the prior slot-domain membership relations and dialogue-aware dynamic slot relations, no matter whether the domains are seen or not.

\section{Dynamic Schema Graph Fusion Network}

%We incorporate dynamic schema graph to fill the slots. 
%The intuition behind the fusion of the static and dynamic schema graph to find the values of the slots is
%The intuition is 
%that strong connections among slots exploit complementary and more global information to the token-level representation created from the given dialogue history, and help boost the accuracy of dialogue state tracking. 
%In particular, the strong connections do not only rely on prior schema knowledge, but also on the dialogue history. Therefore, we employ dynamic completion strategies that determine at each turn which missing relations between slots for specific dialogue context should be added to the initial schema graph to better fill slots.

The proposed DSGFNet consists of four components: (1) a \textit{BERT-based dialogue utterance encoder} that aims to contextualize the tokens of the current turn and the dialogue history; (2) a \textit{schema graph encoder} that is able to generalize to unseen domains and shares information among predefined slot-domain membership relations;
% \QZ{is able to generalize to unseen domains and allows information propagation among slots and domains}.~\comment{Qiang}{emphasize what's new.} 
%encodes each slot node based on the predefined static slot relations and its corresponding descriptions;
(3) a \textit{dialogue-aware schema graph evolving network} that adds the dialogue-aware dynamic slot relations into the schema graph; and (4) a \textit{schema graph enhanced dialogue state decoder} that extracts the value span from the candidate elements based on the evolved schema graph. Figure~\ref{fig:framework} illustrates the architecture.

%We encode the dialogue context with  BERT, and model the information interactions among the schema graph nodes (the slots) with a graph attention network. The dynamic schema graph completion network utilizes dialogue attention to dynamically augment the relation space of each slot during the dialogue state tracking process. The resulting contextual representations of dialogue tokens and slots are feed-forwarded to various classification heads in the dialogue state decoder to solve the sub-tasks for dialogue state tracking.

\subsection{Dialogue Utterance Encoder}

\renewcommand{\vec}{\bm}
% It can capture the relations between the current utterance and the previous utterances.

This encoder takes as input the current and previous dialogue utterances. Specifically, the input is a sequence of tokens with length $K$, i.e., $[t_1, ..., t_K]$. Here, we set the first token $t_1$ to \texttt{[CLS]}; subsequent are the tokens in the current dialogue utterance and the ones in the previous dialogue utterances, which are separated by \texttt{[SEP]}. We employ BERT~\citep{devlin2018bert} to obtain contextual token embeddings. The output is a tensor of all the token embeddings $\vec{B}=[\vec{b}_1, ..., \vec{b}_K]$,  with one embedding for each token.

\subsection{{Schema Graph Encoder}}
\label{section:graph}

To make use of the slot-domain membership relations defined by prior domain knowledge, we construct a schema graph based on the predefined ontology. An example is shown in Figure~\ref{fig:framework}. In this schema graph, 
each node represents either a domain or a slot,
%Each node is represented by an embedding. %by vectors in a semantic space as nodes initial embeddings. 
and all the slot nodes are connected to their corresponding domain nodes. %by There are edges between slots and the domain which the slots belong to. 
In order to allow information propagation across domains, all the domain nodes are connected with each other. %All domains are connected in order to allow the propagation of information across domains.

\textbf{Schema-Agnostic Embedding Initializer.}
% So as to generalize to unseen domains and slots, 
To generalize to unseen domains, 
% and share knowledge across-domains
DSGFNet initializes the schema graph node embeddings via a schema-agnostic projection.
Inspired by zero-shot learning~\citep{romera2015embarrassingly}, 
% we constrain the schema element semantic encoder not to have any domain- or slot-specific parameters so that it can generalize to unseen schema. 
we propose a schema-agnostic embedding initializer to project schemata across domains into a unified semantic distribution. Specifically, we feed a natural language description of one slot/domain into BERT, using the output of \texttt{[CLS]} as the semantic embeddings for this slot/domain. The semantic embeddings for the set of slot and domain is
$\vec{I} = [\vec{i}_1, ..., \vec{i}_{N+M}]$, where $N$ and $M$ are the number of slots and domains, respectively. We constrain the schema embedding initializer not to have any domain-specific parameters so that it can generalize to unseen domains.
% \comment{Qiang}{Give an equation}
% The schema element semantic encoder has two benefits: 
% 1) since it is trained across domains, it allows the cross-domain sharing of knowledge; and 
% 2) once the encoder is fine-tuned, it can be used to process any unseen domains and slots.

\textbf{Slot-Domain Membership Relation Reasoning Network.}
To involve the prior slot-domain membership relations into the schema graph node embeddings, DSGFNet propagates information among slots and domains over the schema graph. We add a self-loop to each node because the nodes need to propagate information to themselves. %the information from itself in the information propagation process. 
Inspired by the GAT model~\cite{velivckovic2017graph}, we propose a slot-domain membership relation reasoning network to propagate information over the schema graph. For each node, we first compute attention scores $\vec{\alpha}$ for its neighbours. These attention scores are used to weigh the importance of each neighboring node. % different nodes within the neighborhood. 
Formally, the attention scores are calculated as follows:
\begin{gather}
h_{i,j} = %
\text{ReLU}({\bf W}^\top\cdot[\vec{i}_i, \vec{i}_j]),\\ %
\alpha_{i,j} = %
\frac{\text{exp}(h_{i,j})}{\sum_{k \in \set{N}_i}\text{exp}(h_{i,k})},
\end{gather}
where ${\bf W}$ is a matrix of parameters and 
$\set{N}_i$ is the neighborhood of the $i$-th node. 
The normalized attention coefficients and the activation function are used to compute a non-linear weighted combination of the neighbours. This is used to compute the tensor of the schema graph node embeddings $\vec{G} = (\vec{g}_1, ..., \vec{g}_{N+M})$:
\begin{gather}
\vec{g}_i = \text{ReLU} %
\left(\sum_{j \in \mathcal{N}_i} \alpha_{i,j} \cdot \vec{i}_j\right),
\end{gather}
where $ i \in \{1,\dots, N+M\}$. To explore the higher-order connectivity information of slots across domains, we stack $l$ layers of the reasoning network. Each layer takes the node embeddings from the previous layer as input, and outputs the updated node embeddings to the next layer. 
% Slots are represented by the embeddings of the corresponding schema graph node.

\subsection{Schema Graph Evolving Network}

% Dynamic slot relations, such as 
% co-reference, co-occurrence and co-update, 
% depend on the dialogue context. %need to be dynamically decided according to dialogue context. 
% Dynamic extraction of such slot relations is important for understanding dialogue history and enhancing the performance of the DST. %dialogue state tracking. 
We propose a schema graph evolving network to incorporate the dialogue-aware dynamic slot relations into the schema graph, which is composed of two layers, a {schema-dialogue fusion layer} and a {dynamic slot relation completion layer}.

\newcommand{\aware}{\tilde}
% Attention mechanism has the ability to well capture the important dependency among information, therefore,
\textbf{Schema-Dialogue Fusion Layer.} %
Since the dynamic slot relations are related to the dialogue context, we need to fuse the dialogue context information into the schema graph. We adopt the multi-head attention~\citep{vaswani2017attention} to achieve this goal. The mathematical formulation is:
\begin{gather}
\vec{H} = \text{MultiHead}(\text{Q} = \vec{g_i}, \text{K} = \vec{B}, \text{V} = \vec{B}),\\
\aware{\vec{g}_i} = \vec{H} \cdot \vec{W}_a,
\end{gather}
where $\vec{W}_a$ is learnable parameters of a linear projection after the multi-head attention, and $\aware{\vec{g}_i}$ is the dialogue-aware schema graph node embeddings.

\textbf{Dynamic Slot Relation Completion Layer.} %
This layer aims to augment the dynamic slot relations on the schema graph based on the dialogue-aware node embeddings.
To involve the dialogue-aware dynamic slot relations into DST explicitly, DSGFNet defines three types of dynamic slot relations: (1) Co-reference relations occur when a slot value has been mentioned earlier in the dialogue and has been assigned to another slot; (2) Co-update relations occur when slot values are updated together at the same dialogue turn, and; (3) Co-occurrence relations occur when slots with a high co-occurrence probability in a large dialogue corpus appear together in the current dialogue. 
Specifically, we feed the dialogue-aware slot node representations into a multi-layer perceptron followed by a $4$-way softmax function to identify the relations between slot pairs, which include the $none$ relation and the three dynamic relations mentioned above. Formally, given the $i$-th and $j$-th dialogue-aware slot node embeddings $\aware{\vec{g}}_i$ and $\aware{\vec{g}}_j$, we obtain an adjacent matrix of the dynamic slot relations for all slot pairs as follows:
\begin{equation}
\vec{A}(i,j) = \mathop{\text{arg\,max}}\left(\text{softmax}(\text{MLP}(\aware{\vec{g}}_i \oplus \aware{\vec{g}}_j))\right).
\end{equation} 
With $\vec{A}$, we add dynamic slot relation edges to the schema graph.

\subsection{Dialogue State Decoder}
To decode the slot values by means of incorporating the slot-domain membership relations and dialogue-aware dynamic slot relations which are captured by the evolved schema graph, we propose a schema graph enhanced dialogue state decoder. 

% The evolved schema graph contains multiple types of relations. 
To learn a more comprehensive slot node embedding, we need to fuse multiple relations on the evolved schema graph.
% The dialogue state decoder aims to fuse the dynamic schema graph to obtain complementary and global information about the slot relations. Towards this end, %we first need to combine the static and dynamic relations in the schema graph need to be combined firstly, 
DSGFNet divides different relations on the schema graph into sub-graphs $R_s, R_r, R_u, R_o$, which represent slot-domain membership relation, co-reference relation, co-update relation, and co-occurrence relation, respectively. For each sub-graph $R_i$, its node embeddings $\vec{s}_i$ are obtained by attending over the neighbors, which is the same as the method used in Section~\ref{section:graph}. Considering that different relation types have different contributions to the node interactions for different dialogue contexts~\citep{wang2019heterogeneous}, we aggregate these different sub-graphs via an attention mechanism as follows:
\begin{gather}
\vec{S} = [\vec{s}_s;\vec{s}_r;\vec{s}_u;\vec{s}_o], \\
\vec{\beta} = \text{softmax}(\vec{S}^\top\cdot\text{tanh}(\vec{W}_s \cdot \vec{b}_{[CLS]} + \vec{b}_s)),\\
\vec{s}= \vec{S} \cdot \vec{\beta},
\end{gather}
where $\vec{W}_s$, $\vec{b}_s$ are learnable weights, $\vec{b}_{[CLS]}$ is the output of BERT-based dialogue utterance encoder. 

Each slot value is extracted by a value predictor based on the corresponding fused slot node embeddings $\vec{s}$. The value predictor is a trainable nonlinear classifier followed by two parallel softmax layers to predict start and end positions in candidate elements $\vec{C}$, which are composed by the dialogue context $\vec{B}$ and slots' candidate value vocabulary $\vec{V}$:
\begin{gather}
\vec{C} = [\vec{B}; \vec{V}] \\
[\vec{l}_s, \vec{l}_e] = \vec{r}_d \cdot \text{tanh}(\vec{s}^\top \cdot \vec{W}_d \cdot \vec{C} + \vec{b}_d), \\
p_s = \text{softmax}(\vec{l}_s), \\
p_e = \text{softmax}(\vec{l}_e),
\end{gather}
where $\vec{r}_d$, $\vec{W}_d$, and $\vec{b}_d$ are trainable parameters. Note that if the end position is before the start position, the resulting span will simply be ``None". If the start position is in the slots’ candidate value vocabulary, the resulting span will only pick the candidate value in this position. 
% The extracted dialogue span is the predicted value for slot.

\subsection{Training and Inference}
During training, we use ground truth dynamic slot relation graph to optimize the dialogue state decoder. Cross-entropy between predicted value span $[p_s, p_e]$ and ground truth value span is utilized to measure the loss of the value span prediction $\mathcal L_s$.  
The dynamic slot relation identifier is optimized by the cross-entropy loss $\mathcal L_r$ between predicted dynamic relation $\vec{A}$ and the ground truth dynamic slot relation. 
We train dialogue state decoder and dynamic slot relation identifier together, the joint loss $\mathcal L$ is computed as follows:
\begin{gather}
\mathcal L = \lambda \cdot \mathcal L_r + (1-\lambda) \cdot\mathcal L_s,
\end{gather}
where $\lambda \in [0, 1]$ is a balance coefficient.
During inference, the predicted dynamic slot relation $\vec{A}$ is used to predict value span as dialogue state.

\section{Experiments}

\subsection{Datasets}

We conduct experiments on three task-oriented dialogue benchmark datasets: %
SGD~\citep{rastogi2019towards}, %
MultiWOZ2.2~\citep{zang2020multiwoz}, and %
MultiWOZ2.1~\citep{eric2019multiwoz}. %
Among them, SGD is by far the most challenging dataset which contains over 16,000 conversations between a human-user and a virtual assistant across 16 domains. Unlike the other two datasets, it also includes unseen domains in the test set. %
MultiWOZ2.2 and MultiWOZ2.1 are smaller human-human conversations benchmark datasets, which contain over 8,000 multi-turn dialogues across 8 and 7 domains, respectively. MultiWOZ2.2 is a revised version of MultiWOZ2.1, which is re-annotated with a different set of annotators and also canonicalized entity names. 
% DSGFNet uses span prediction to predict the dialogue state, therefore we follow the same data preprocessing in TripPy~\citep{heck2020trippy} to make every slot-value pointable. 
Details of datasets are provided in Table~\ref{tab:datasets}.

\begin{table}[h]
\centering
\caption{Characteristics of the datasets in experiments. The numbers provided are for the training sets of the corresponding datasets.} 
% \vspace{-0.5em}
\resizebox{0.48\textwidth}{!}{%
\begin{tabular}{l|ccc}
        \toprule
        \bf{Characteristics}&\bf{SGD}&\bf{MultiWOZ2.2} &\bf{MultiWOZ2.1} \\
 		\hline
        \hline
        \text{No. of domains} & 16 & 8 & 7 \\
        \text{No. of dialogues} & 16,142 & 8,438 & 8,438\\
        \text{Total no. of turns} & 329,964 & 113,556&  113,556\\
        \text{Avg. turns per dialogue} & 20.44 & 13.46&  13.46\\
        \text{Avg. tokens per turn} & 9.75 & 13.13&  13.38\\
        \text{No. of slots} & 215 & 61 &  37\\
        % \text{No. of categorical slots} & 53 & 21 &  \\
        % \text{No. of non-categorical slots} & 162 & 40 &  \\
% 		\textit{No. of slot values} & 14,139 && 212 & 99 & 138 &\\
% 		\text{Schema description} & Yes & Yes &   \\
		\text{Unseen domains in test set} & Yes &No & No\\
		\bottomrule
	\end{tabular}
	}
% \vspace{-1em}
\label{tab:datasets}
\end{table}

\subsection{Baselines}
We compare with the following existing models, which are divided into two categories.
% predicting the dialogue state independent of the relations among domains and slots, or based on such relations. 
(1) Models that can predict dialogue state on unseen domains:
~\textit{SGD-baseline}~\citep{rastogi2019towards}, a schema-guided paradigm that predicts states for unseen domains;
~\textit{FastSGT}~\citep{noroozi2020fast}, a BERT-based model that uses multi-head attention projections to analyze dialogue;
~\textit{Seq2Seq-DU}~\citep{feng2020sequence}, a sequence-to-sequence framework which decodes dialogue states in a flatten format.
(2) Models that cannot predict dialogue state on unseen domains:
~\textit{TRADE}~\citep{wu2019transferable}, a generation model which generates dialogue states from utterances using a copy mechanism; ~\textit{DS-DST}~\citep{zhang2019find}, a dual strategy that classifies over a picklist or finding values from a slot span; 
~\textit{TripPy}~\citep{heck2020trippy}, an open-vocabulary model which copies values from dialogue context, or slot values in previous dialogue state;
~\textit{SOM-DST}~\citep{kim2020efficient}, a selectively overwriting mechanism which first predicts state operation on each of the slots and then overwrites with new values; ~\textit{MinTL-BART}~\citep{lin2020mintl}, a plug-and-play pre-trained model which jointly learns dialogue state tracking and dialogue response generation; 
~\textit{SST}~\citep{chen2020schema}, a graph model which fuses information from utterances and static schema graph;
~\textit{PPTOD}~\citep{su2021multi}, a multi-task pre-training strategy that allows the model to learn the primary TOD task completion skills from heterogeneous dialog corpora.

\subsection{Evaluation Measures}
Our evaluation metrics are consistent with prior
works on these datasets. We compute the Joint Goal Accuracy (Joint GA) on all test sets for straightforward comparison with the state-of-the-art methods. Joint GA is defined as the ratio of dialogue turns for which all slots have been filled with the correct values according to the ground truth. 

\subsection{Training}
We use BERT model (i.e., BERT-base and uncased) to encode utterances and schema descriptions. The BERT models are fine-tuned in the training process. The maximum length of an input sequence is set to 512. The hidden size of the schema graph encoder and the schema graph evolving network is set to 256. The dropout probability is 0.3. The balance coefficient $\lambda$ is 0.5. Adam~\citep{kingma2014adam} is used for optimization with an initial learning rate (LR) of 2e-5. We conduct training with a warm-up proportion of 10\% and let the LR decay linearly after the warm-up phase. 
% Hyper parameters are tuned and chosen based on the validation dataset in all cases. 
% The effects of some crucial parameters are shown in Appendix~\ref{sec:appendix}.

%\subsection{Experimental Results}
\section{Results and Discussion}
Tables~\ref{tab:SGD}, ~\ref{tab:MultiWOZ2.2}, ~\ref{tab:MultiWOZ2.1} show the performance of DSGFNet as well as the baselines on three datasets respectively. It is shown that DSGFNet achieves state-of-the-art performance in unseen domains on SGD, all domains on SGD, and MultiWOZ2.2. All improvements observed compared to the baselines are statistically significant according to two sided paired t-test (p < 0.05). And the performance on MultiWOZ2.1 are comparable with the state-of-the-art\footnote{TRADE, SST use the original MultiWOZ datasets. The other models use the data preprocessed by TripPy.}. Most notably, DSGFNet improves the performance on SGD most significantly, which has unseen domains and more complex schemata domains, compared to the runner-up. 
%The improvements on the smaller datasets MultiWOZ2.1 and MultiWOZ2.2 demonstrate that our approach benefits from its design on dynamic schema graph as well. 
% This demonstrates the success of the dynamic schema graph in DSGFNet.
% the success of DSGFNet is due to its suitable architecture of a dynamic schema graph. 
It indicates that DSGFNet can facilitate knowledge transfer to new domains and improve relation construction among complex schemata domains. We conjecture that it is due to DSGFNet containing the schema-agnostic encoder and dynamic schema graph.
% The more plentiful the relations among domains and slots are, the better performance DSGFNet can achieve. 
The following analysis provides a better understanding of our model’s strengths.

\begin{table}[!h]
\centering
\caption{Joint GA of DSGFNet and baselines in unseen domains and all domains on SGD dataset. DSGFNet significantly improves over the best baseline (two-sided paired t-test, $p<0.05$).}
% \vspace{-0.5em}
\resizebox{0.5\textwidth}{!}{
\begin{tabular}{l|c|c}
        \toprule
        \bf{Models} & \makecell[c]{\bf{SGD} \\\bf{Unseen Domains}} & \makecell[c]{\bf{SGD} \\\bf{All Domains}} \\
        % \cline{2-3}
        % &{Joint GA} &{Joint GA}  \\
 		\hline
        \hline
        \text{SGD-baseline}~\citep{rastogi2019towards} & 20.0\% &25.4\% \\
        \text{FastSGT}~\citep{noroozi2020fast} & 20.8\% & 29.2\% \\
        \text{Seq2Seq-DU}~\citep{feng2020sequence} & 23.5\% & 30.1\%\\
        \hline
		\bf{\text{DSGFNet} } & \bf{24.4\%} & \bf{32.1\%}\\
		\toprule
	\end{tabular}
} 
% \vspace{-1.5em}
\label{tab:SGD}
\end{table}

\begin{table}[!h]
\centering
\caption{Joint GA of DSGFNet and baselines on MultiWOZ2.2. DSGFNet significantly improves over the best baseline (two-sided paired t-test, $p<0.05$).}  
% \vspace{-0.5em}
\resizebox{0.45\textwidth}{!}{
\begin{tabular}{l|c}
        \toprule
        \bf{Model} & \bf{MultiWOZ2.2} \\
        % \cline{2-3}
        % &{Joint GA} &{Joint GA}  \\
 		\hline
        \hline
        \text{SGD-baseline}~\citep{rastogi2019towards} &42.0\%\\
        \text{TRADE}~\citep{wu2019transferable} &45.4\% \\
        \text{DS-DST}~\citep{zhang2019find} & 51.7\% \\
        \text{TripPy}~\citep{heck2020trippy} &53.5\% \\
        \text{Seq2Seq-DU}~\citep{feng2020sequence} & 54.4\%\\
        \hline
		\bf{\text{DSGFNet}} & \bf{55.8\%  } \\
		\toprule
	\end{tabular}
}
% \vspace{-1.5em}
\label{tab:MultiWOZ2.2}
\end{table}

\begin{table}[!h]
\centering
\caption{Joint GA of DSGFNet and baselines on MultiWOZ2.1. DSGFNet achieves comparable performance of the best baseline.}  
% \vspace{-0.5em}
\resizebox{0.45\textwidth}{!}{
\begin{tabular}{l|c}
        \toprule
        \bf{Model} & \bf{MultiWOZ2.1} \\
        % \cline{2-3}
        % &{Joint GA} &{Joint GA}  \\
 		\hline
        \hline
        \text{SGD-baseline}~\citep{rastogi2019towards} & 43.4\%\\
        \text{TRADE}~\citep{wu2019transferable} & 46.0\%\\
        \text{DS-DST}~\citep{zhang2019find} & 51.2\%\\
        % \text{LABES-S2S}~\citep{zhang2020probabilistic} &51.5\%\\
        \text{SOM-DST}~\citep{kim2020efficient} &53.0\%\\
        \text{MinTL-BART}~\citep{lin2020mintl} & 53.6\%\\
        \text{SST}~\citep{chen2020schema} & 55.2\%\\
        \text{TripPy}~\citep{heck2020trippy} & 55.3\%\\
        \text{PPTOD}~\citep{su2021multi} & \bf{57.1\%}\\
        \hline
		\bf{\text{DSGFNet}} & 56.7\% \\
		\toprule
	\end{tabular}
}
% \vspace{-1.5em}
\label{tab:MultiWOZ2.1}
\end{table}

\begin{table}[!h]
\centering
\caption{Ablation study on unseen domains of SGD, all domains of SGD, MultiWOZ2.2 and MultiWOZ2.1.}  
% \vspace{-0.5em}
\resizebox{0.49\textwidth}{!}{
\begin{tabular}{l|c|c|c|c}
        \toprule
        \makecell[c]{\bf{Model}} & \makecell[c]{\bf{Joint GA} \\ \bf{Unseen}\\\bf{ Domains}\\\bf{SGD}} &\makecell[c]{\bf{Joint GA} \\ \bf{All}\\\bf{ Domains}\\\bf{SGD}} & \makecell[c]{\bf{Joint GA} \\\bf{MultiWOZ} \\\bf{2.2}} &\makecell[c]{ \bf{Joint GA}\\ \bf{MultiWOZ} \\\bf{2.1}}\\
        % \cline{2-3}
        % &{Joint GA} &{Joint GA}  \\
 		\hline
        \hline
        \text{\bf{DSGFNet}}&24.4\%& 32.1\% & 55.8\% & 56.7\%\\
        \text{-w/o Slot-Domain Membership Relations}&21.9\%& 29.8\% & 53.4\% & 54.1\% \\
        \text{-w/o Dynamic Slot Relations}&20.6\%& 28.6\% & 52.2\% & 53.2\% \\
        \text{-w/o Relation Aggregation}&23.8\%& 31.5\% & 55.2\% & 55.9\% \\
		\toprule
	\end{tabular}
}
% \vspace{-1em}
\label{tab:ablation}
\end{table}

\subsection{Ablation Study}

We conduct an ablation study on DSGFNet to quantify the contributions of various factors: the usage of slot-domain membership relations, dynamic slot relations, and multiple relation aggregation. The results indicate that the dynamic schema graph of DSGFNet is indispensable for DST.

\subsubsection*{Effect of Slot-Domain Membership Relations}
To check the effectiveness of the slot-domain membership relations, we remove the schema graph by replacing the prior slot-domain relation adjacency matrix with an identity matrix ${\bm{I}}$. Results in Table~\ref{tab:ablation} show that the joint goal accuracy of DSGFNet without the slot-domain membership relations decreases markedly on unseen domains of SGD, all domains of SGD, MultiWOZ2.2, and MultiWOZ2.1. It indicates the schema graph, which contains slot-domain membership relations, can facilitate knowledge sharing among domain and slot no matter whether the domain is seen or not.

\subsubsection*{Effect of Dynamic Slot Relations}
To investigate the effectiveness of the dialogue-aware dynamic slot relations in the schema graph, we eliminate the evolving network of DSGFNet. Table~\ref{tab:ablation} shows the results on unseen domains of SGD, all domains of SGD, MultiWOZ2.2, and MultiWOZ2.1 in terms of joint goal accuracy. One can observe that without the dynamic slot relations the performance deteriorates considerably. In addition, there is a more markedly performance degradation compared with the results of the slot-domain membership relations. It indicates that the dynamic slot relations are more essential for DST, which can facilitate the understanding of the dialogue context.
% Moreover, in SGD which contains more complex schema, there is a more significant performance degradation compared with the results in MultiWOZ2.2 and MultiWOZ2.1. It indicates that the dynamically-evolving schema graph can make more improvements over complicated domains.

\subsubsection*{Effect of Multiple Relation Aggregation}
To validate the effectiveness of the schema graph relation aggregation mechanism in the dialogue state decoder, we directly concatenate all sub-graph representations instead of calculating a weighted sum via the sub-graph attention. As shown in Table~\ref{tab:ablation}, the performance of the models without the relation aggregation layer in terms of joint goal accuracy decreases markedly compared to DSGFNet. It indicates that the attentions to different types of relations affect the dialogue understanding ability.

\subsection{Further Analysis}

\subsubsection*{Prediction of Dynamic Slot Relations}
\begin{figure}[!t]
\centering
\includegraphics[width=0.48\textwidth]{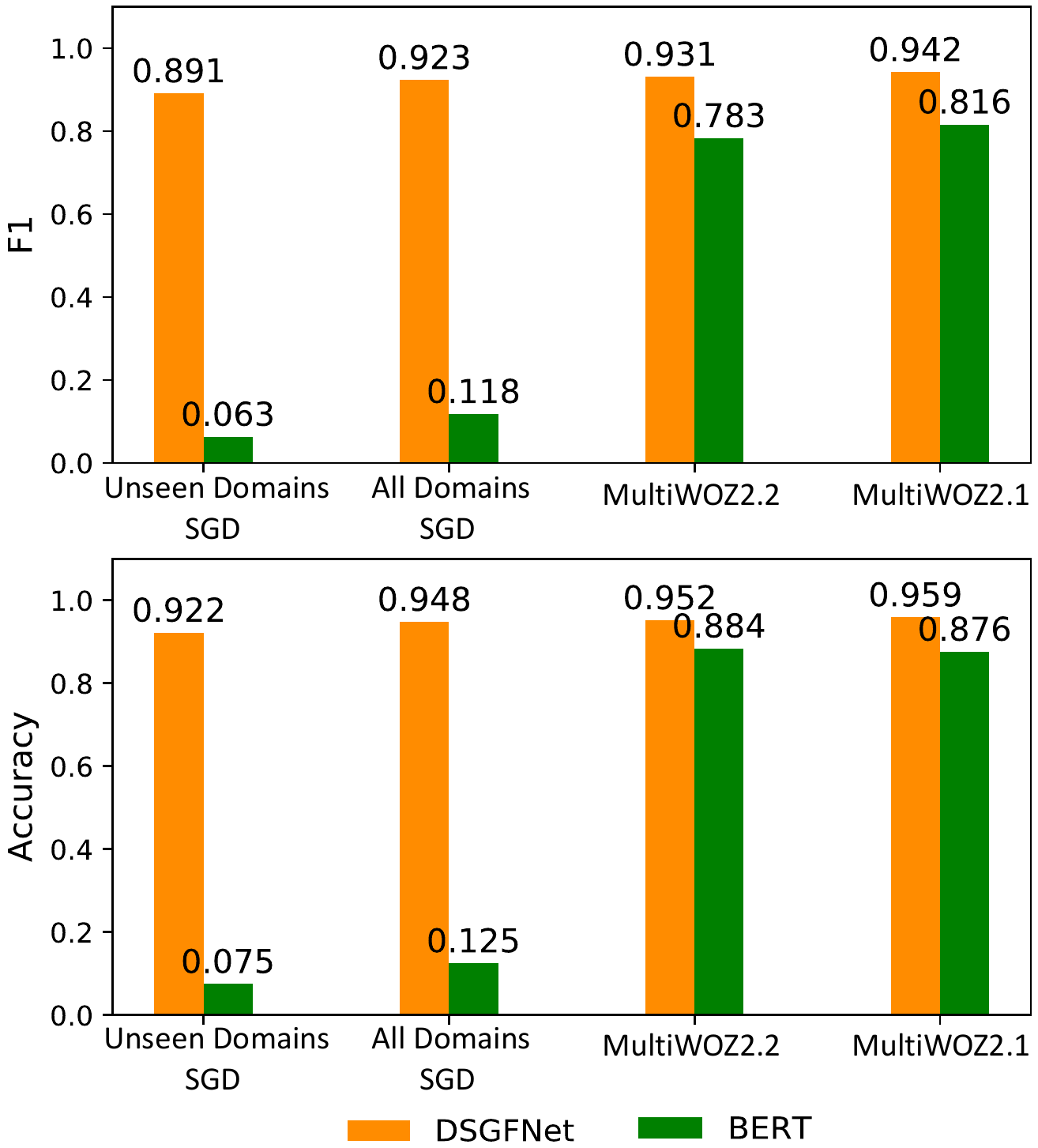}
% \vspace{-0.5em}
\caption{F1 and Accuracy of DSGFNet and BERT for dynamic relation prediction on unseen domains SGD, all domains of SGD, MultiWOZ2.2 and MultiWOZ2.1.}
\label{fig:classification}
% \vspace{-1em}
\end{figure}

In order to test the discriminative capability of DSGFNet for dynamic slot relations, we evaluate the performance of the schema graph evolving network. Since baselines cannot predict the dynamic slot relations explicitly, we compare DSGFNet with the BERT-based classification approach. Following the classification task in BERT, the input sequence starts with [CLS], followed by the tokens of the dialogue context and slot pairs, separated by [SEP], and the [CLS] representation is fed into an output layer for classification. Figure~\ref{fig:classification} shows the results on unseen domains of SGD, all domains of SGD, MultiWOZ2.2, and MultiWOZ2.1 in terms of F1 and Accuracy. From the results, we observe that DSGFNet outperforms BERT significantly. We conjecture that it is due to the exploitation of schema graph with slot-domain membership relations in DSGFNet. In addition, since BERT without schema encoder cannot solve unseen domains, there is a significant performance degradation on SGD which contains a large number of unseen domains in the test set.

\begin{table*}[!ht]
\centering
\caption{Case study of DSGFNet and Seq2Seq-DU on SGD. Slot values are extracted from the dialogue context with the same color. The relation of yellow high-light slot pair is predicted as co-reference. The relation of red underline slot pair is predicted as co-update. The relation of bold font slot pair is predicted as co-occurrence. Slot values in red high-light are incorrectly predicted ones.}
% \vspace{-5pt}
\resizebox{1.0\textwidth}{!}{%
\begin{tabular}{l||l}
        \toprule
        \bf{Dialogue Utterance}& 
        \makecell[l]{\textit{$[$User$]$}: What's the weather going to be like in {\color{red}vancouver} on {\color{yellow}March 10th}? \\
        \textit{$[$Sys$]$}: The forecast average is 68 degrees with a 25 per cent chance of rain. \\
        \textit{$[$User$]$}: Any good attractions in {\color{cyan}town}? \\
        \textit{$[$Sys$]$}: I have 10 good options including {\color{green}Bloedel Conservatory}, a city park. \\
        \textit{$[$User$]$}: Lovely! Can you book me a ride there? \\
        \textit{$[$Sys$]$}: Do you want a luxury or pool ride? How many people? \\
        \textit{$[$User$]$}: Just a {\color{magenta}regular} ride please, book for {\color{blue}1}. \\
        \textit{$[$Sys$]$}: Confirming you want to book a regular cab to Bloedel Conservatory for 1 person.} \\
        \hline
        \bf{Ground Truth Dialogue State}& 
        \makecell[l]{\textit{{$[$Weather$]$}}: city = ``{\color{red}vancouver}"; date = ``{\color{yellow}March 10th}"; \\
        \textit{$[$Travel$]$}: location = ``{\color{cyan}vancouver}";\\
        \textit{$[$RideSharing$]$}: destination = ``{\color{green}Bloedel Conservatory}"; number of seats = ``{\color{blue}1}"; ride type = ``{\color{magenta}regular}";}\\
        \hline
        \bf{State Predictions of DSGFNet}& 
        \makecell[l]{\textit{{$[$Weather$]$}}: \colorbox{yellow}{\reduline{{\bf city}}} = ``vancouver"; \reduline{{\bf date }}= ``March 10th"; \\
        \textit{$[$Travel$]$}: \colorbox{yellow}{location} = ``vancouver";\\
        \textit{$[$RideSharing$]$}: {destination} = ``Bloedel Conservatory"; \reduline{number of seats} = ``1"; \reduline{ride type} = ``regular";} \\
        \hline
        % \makecell[l]{\bf{State Predictions of DSGFNet \\ \bf{(-w/o Dynamically Evolving)}}}&
        % \makecell[l]{\textit{{$[$Weather$]$}}: city = ``vancouver"; date = ``March 10th"; \\
        % \textit{$[$RideSharing$]$}: destination = ``Bloedel Conservatory"; number of seats = ``1";} \\
        % \hline
        % \makecell[l]{\bf{State Predictions of DSGFNet} \\ \bf{(-w/o Schema Graph)}}& 
        % \makecell[l]{\textit{{$[$Weather$]$}}: city = ``vancouver"; date = ``March 10th"; \\
        % \textit{$[$RideSharing$]$}: destination = ``Bloedel Conservatory"; }\\
        % \hline
        \bf{State Predictions of Seq2seq-DU}& 
        \makecell[l]{\textit{{$[$Weather$]$}}: city = ``vancouver"; date = ``March 10th"; \\
        \textit{$[$Travel$]$}: location= \colorbox{red}{``town"};\\
        \textit{$[$RideSharing$]$}: {destination} = ``Bloedel Conservatory"; number of seats = ``1"; ride type = \colorbox{red}{none};} \\
        % \makecell[l]{\textit{{$[$Weather$]$}}: city = ``vancouver"; date = ``March 10th"; \\
        % \textit{$[$Travel$]$}: location = ``vancouver";\\
        % \textit{$[$RideSharing$]$}: destination = ``Bloedel Conservatory"; number of seats = ``1"; ride type = ``regular";} \\
 		\toprule
	\end{tabular}
	}
\vspace{-1em}
\label{tab:error}
\end{table*}

\begin{table}[!ht]
\centering
\caption{Performance comparison with different dynamic slot relations and fully-connected relations on unseen domains of SGD, all domains of SGD, MultiWOZ2.2 and MultiWOZ2.1.}  
% \vspace{-0.5em}
\resizebox{0.49\textwidth}{!}{
\begin{tabular}{l|c|c|c|c}
        \toprule
        \makecell[c]{\bf{Model}} & \makecell[c]{\bf{Joint GA} \\\bf{Unseen} \\\bf{Domains} \\\bf{SGD}} & \makecell[c]{\bf{Joint GA} \\\bf{All} \\\bf{Domains} \\\bf{SGD}} & \makecell[c]{\bf{Joint GA} \\\bf{MultiWOZ} \\\bf{2.2}} &\makecell[c]{ \bf{Joint GA}\\ \bf{MultiWOZ} \\\bf{2.1}}\\
        % \cline{2-3}
        % &{Joint GA} &{Joint GA}  \\
 		\hline
        \hline
        \text{-w All Dynamic Relations}& 24.4\%& 32.1\% & 55.8\% & 56.7\%\\
        \text{-w Co-reference Relation}& 21.5\%& 29.8\% & 53.9\% & 54.7\% \\
        \text{-w Co-occurrence Relation}& 23.8\%& 31.7\% & 55.3\% & 55.9\% \\
        \text{-w Co-update Relation}& 22.3\%& 30.1\% & 53.5\% & 54.5\% \\
        \text{-w/o Dynamic Relations}& 20.6\%& 28.6\% & 52.2\% & 53.2\% \\
        \text{-w Fully-connected Relations}& 21.3\%& 29.9\% & 54.2\% & 54.9\% \\
		\toprule
	\end{tabular}
}
% \vspace{-1.5em}
\label{tab:dynamic}
\end{table}

\subsubsection*{Effects of Each Type of Dynamic Slot Relation}
To better illustrate the effectiveness of augmenting slot relations on the schema graph, we study how different dynamic slot relations affect the DST performance. Table~\ref{tab:dynamic} presents the joint goal accuracy of DSGFNet with different dynamic relations on unseen domains of SGD, alll domains of SGD, MultiWOZ2.2, and MultiWOZ2.1. One can see that the performance of DSGFNet with each type of dynamic slot relation surpasses that without any dynamic slot relations considerably. Thus, all types of dynamic slot relations in the schema graph are helpful for dialogue understanding. Furthermore, the performance of DSGFNet with co-occurrence relation is superior to the performance with the other two dynamic slot relations. We conjecture that it is due to the fact that a large percentage of dynamic relations is the co-occurrence relation, which has an incredible effect on DST.

\subsubsection*{Effect of Automatic Relation Completion}
To demonstrate the effectiveness of automatically completing each type of slot relations on the schema graphs, we replace four automatically-completed sub-graphs in DSGFNet with four fully-connected graphs. As shown in Table~\ref{tab:dynamic}, the performance of the model with the fully-connected graphs in terms of joint goal accuracy decreases significantly compared to DSGFNet (two-sided paired t-test, $p < 0.05$). We believe that this is caused by the noise introduced by the redundancy captured by the relations between all pairs of slots. In addition, sampling the relations using our strategy can also reduce the memory requirements when the number of slots and domains are large.

\subsubsection*{Case Study}
We make qualitative analysis on the results of DSGFNet and Seq2seq-DU on SGD. We find that DSGFNet can make a more accurate inference of dialogue states by using the dynamic schema graph. For example, as shown in  Table~\ref{tab:error}, ``city"-``location" is predicted as co-reference relation, ``city"-``date" and ``number of seats"-``ride type" are predicted as co-update relation, ``city"-``date" is predicted as co-occurrence relation. Based on the dynamic schema graph, DSGFNet propagates information involving slot-domain membership relations and dynamic slot relations. Thus, it infers slot values more correctly. In contrast, since Seq2seq-DU ignores the dynamic slot relations, it cannot properly infer the values of ``location" and ``ride type", which have dynamic slot relations with other slots.

\section{Conclusion}
We have proposed a new approach to DST, referred to as DSGFNet, which effectively fuses prior slot-domain membership relations and dialogue-aware dynamic slot relations on the schema graph. 
% To improve the explainability of dynamic slot relations, DSGFNet explicitly identifies co-reference, co-update, and co-occurrence relations. 
To incorporate the dialogue-aware dynamic slot relations into DST explicitly, DSGFNet identifies co-reference, co-update, and co-occurrence relations. 
To improve the generalization ability, DSGFNet employs a schema-agnostic graph attention network to share information. Experimental results show that DSGFNet outperforms the existing methods in DST on three benchmark datasets, including unseen domains of SGD, all domains of SGD, MultiWOZ2.1, and MultiWOZ2.2. 
For future work, we intend %to investigate %two directions 
to further enhance our approach by utilizing more complex schemata and data augmentation techniques.
% and generating values out of the domain ontology,  to further enhance the approach.
% , and data augmentation.

\section{Acknowledgments}
This project was funded by the EPSRC Fellowship titled “Task Based Information Retrieval” and grant reference number EP/P024289/1.

\bibliographystyle{acl_natbib}
\bibliography{emnlp2020}

% \newpage
\clearpage

% \appendix

\appendix

\section{Dynamic Slot Relations Label Collection}
\label{sec:appendix}

Dynamic schema graph in DSGFNet has three types of dynamic slot relations, which includes co-reference relations, co-update relations and co-occurrence relations. The labels of these relations are used to train the schema graph evolving network.
We first collected all possible slot pairs from ground truth dialogue state for each dialogue turn. And then we labeled these slot pairs by the rules as follows: 
% (1) The labels of prior slot-domain membership relations are obtained directly from the SGD, MultiWOZ2.1, and MultiWOZ2.2 datasets. These datasets provide the labels that the slots are belonging to which domain. 
(1) If one slot value has been assigned to another slot in earlier turn of the dialogue, we label the relation between these two slots as co-reference. 
(2) If values of two slot in the same dialogue turn are updated together, we label the relation between these two slots as co-update. 
(3) If the co-occurrence probability of two slots in the training set of SGD, MultiWOZ2.1, and MultiWOZ2.2 is higher than 5\%, we label the relation between these two slots as co-occurrence. Table~\ref{tab:relation} shows the  the proportion of different types of dynamic slot relations on these datasets.

\begin{table}[h]
\centering
\caption{The proportion of different types of dynamic slot relations on SGD, MultiWOZ2.2, and MultiWOZ2.1 in training sets.} 
% \vspace{-0.5em}
\resizebox{0.49\textwidth}{!}{%
\begin{tabular}{l|ccc}
        \toprule
        \bf{Relation}&\bf{SGD}&\bf{MultiWOZ2.2} &\bf{MultiWOZ2.1} \\
 		\hline
        \hline
        \text{Co-reference} & 5.11\% & 4.21\% & 4.29\%  \\
        \text{Co-update} & 9.31\% & 4.01\% & 4.13\%\\
        \text{Co-occurrence} & 31.13\% & 37.53\% &  36.53\%\\
        \bottomrule
	\end{tabular}}
% \vspace{-1em}
\label{tab:relation}
\end{table}

\begin{table}[!h]
\centering\caption{Accuracy of DSGFNet in each domain on SGD test set. Domains marked with ‘*’ are those for which the schemata in the test set are not present in the training set. Domains marked with ‘**’ have both the unseen and seen schemata. For other domains, the schemata in the test set are also seen in the training set.} 
% \vspace{-0.5em}
\resizebox{0.49\textwidth}{!}{
\begin{tabular}{l|c||l|c}
        \toprule
        \textbf{Domain} & \textbf{Joint GA}& \textbf{Domain} & \textbf{Joint GA}\\
        \hline
        \hline
        \textit{RentalCars*} & 5.11\% & \textit{Homes} & 22.46\% \\
        \textit{Messaging*} & 5.48\% & \textit{Events*} & 32.02\% \\
        \textit{Payment*} & 7.31\%  & \textit{Hotels**} & 33.13\%  \\
        \textit{Music*} & 11.87\%  & \textit{Movies**} & 42.13\% \\
		\textit{Buses*} & 12.72\% & \textit{Services**} & 45.39\%  \\
		\textit{Trains*} & 16.39\%  & \textit{Travel} & 48.30\% \\
		\textit{Flights*} & 16.64\% & \textit{Alarm*} & 53.27\%  \\
		\textit{Restaurants*} & 17.01\%  & \textit{RideSharing} & 56.42\%  \\
		\textit{Media*} & 20.83\% & \textit{Weather} & 68.49\%  \\
		\toprule
	\end{tabular}
}
% \vspace{-23pt}
\label{tab:domain_performance_sgd}
\end{table}

\section{Performance on Different Domains}
\label{sec:appendix}
We further investigate the performance of DSGFNet on different domains. Table~\ref{tab:domain_performance_sgd} shows the joint goal accuracy of DSGFNet in different domains on SGD. We observe that the presence of schemata in the training data is the major factor affecting the performance. We see that the best performance can be obtained in the domains with all seen schemata. The domains which have partially unseen schemata achieve higher accuracy, such as ``Hotels”, ``Movies”, and ``Services” domains. The accuracy declines in the domains with only unseen schemata, such as ``RentalCars" and ``Messaging". However, among the domains with only unseen schemata, those have similar schemata to training data resulting in superior performance, such as ``Alarm" and ``Events" domains. We conclude that DSGFNet is able to perform zero-shot learning and share knowledge across domains. However, more sharing of information should be utilized to enhance the generalization ability.

\section{Analysis of Parameters in DSGFNet}
\label{sec:appendix}

\begin{figure}[!h]
\centering
\includegraphics[width=0.47\textwidth]{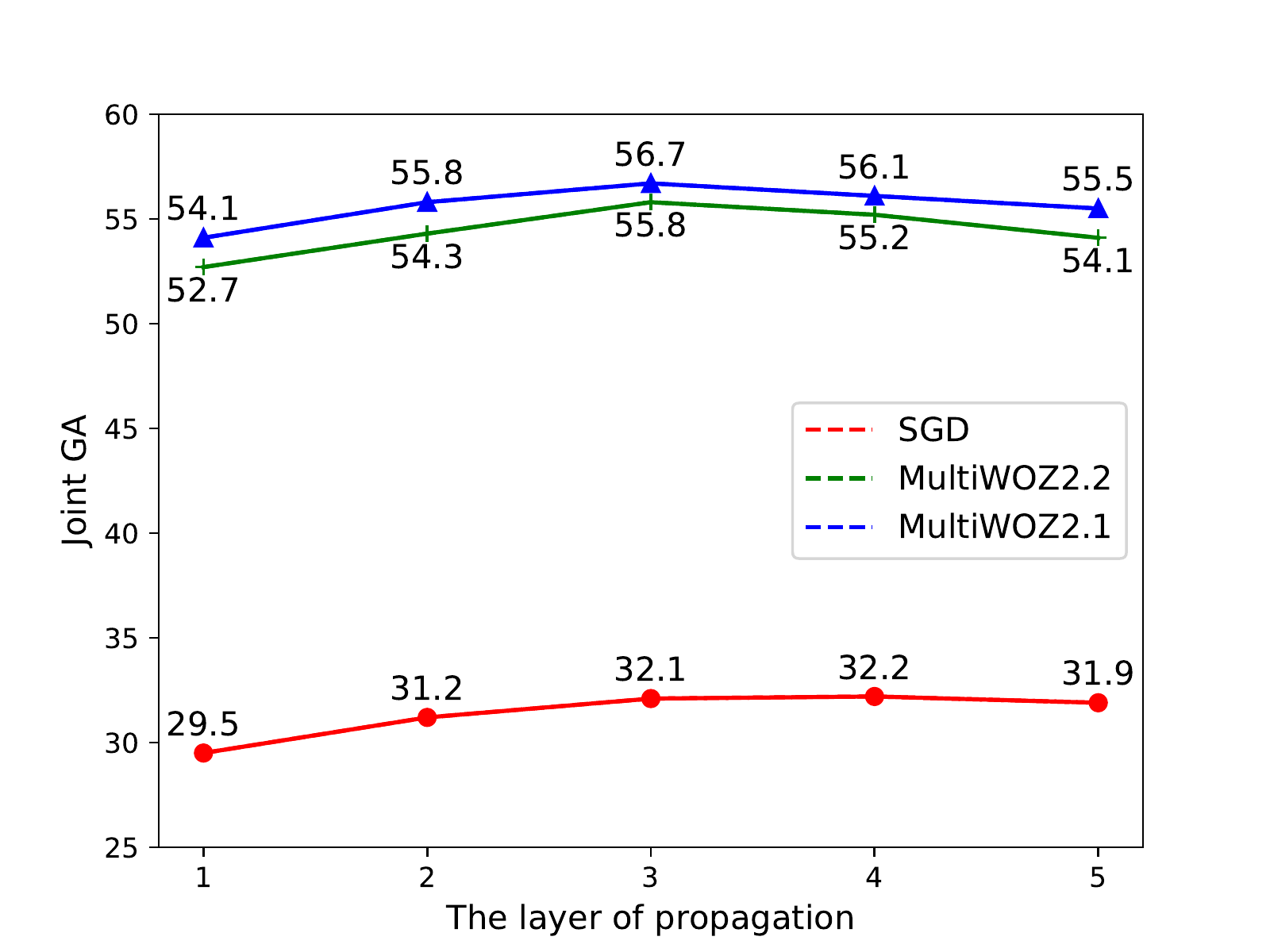}
\caption{Performance comparison \textit{w.r.t.} the layer of propagation on the schema graph.}
\label{fig:parameter1}
\end{figure}

\begin{figure}[!h]
\centering
\includegraphics[width=0.47\textwidth]{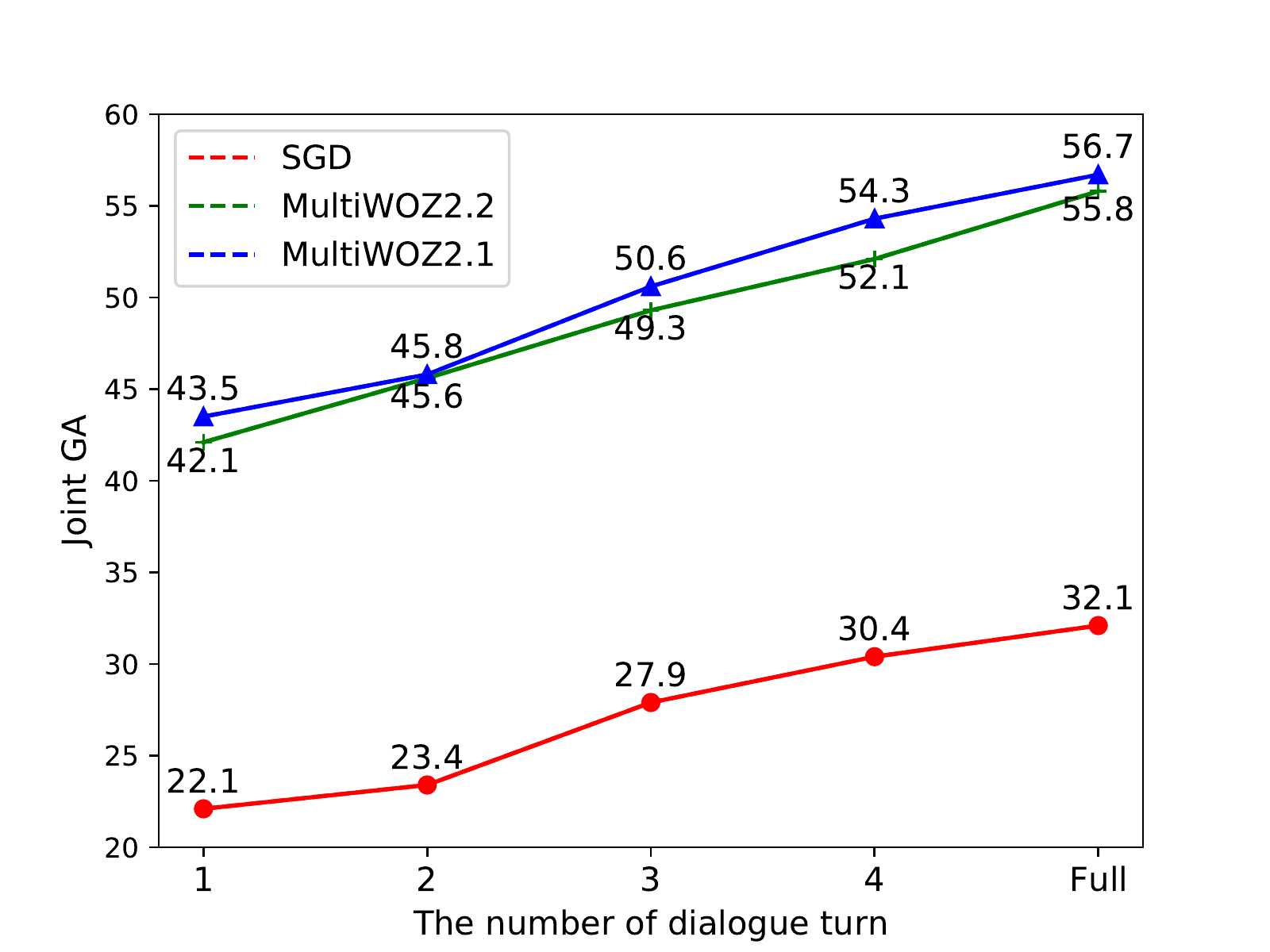}
\caption{Performance comparison \textit{w.r.t.} the number of dialogue turns used in the schema-dialogue fusion layer.}
\label{fig:parameter2}
\end{figure}

\begin{figure}[!h]
\centering
\includegraphics[width=0.47\textwidth]{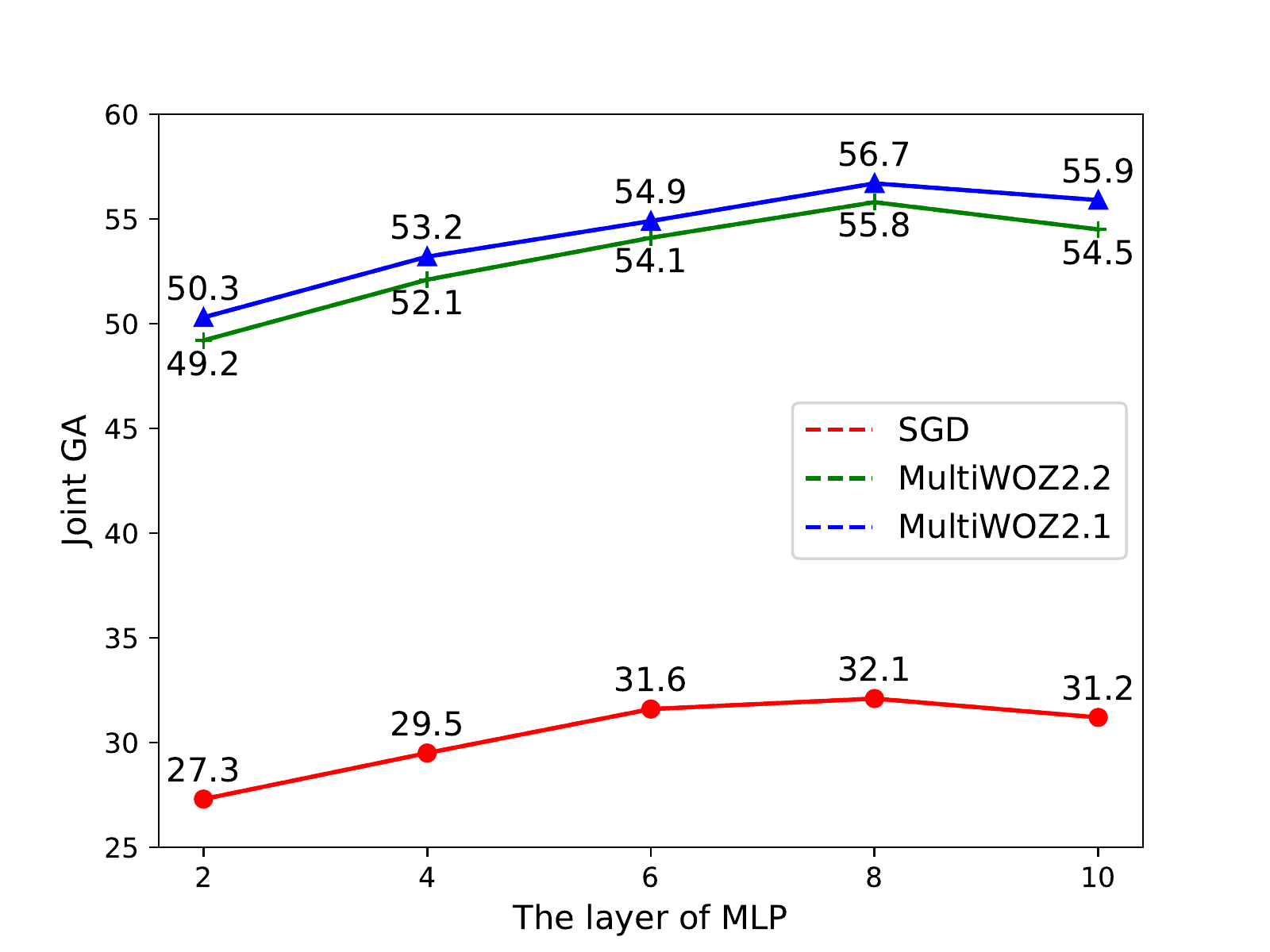}
\caption{Performance comparison \textit{w.r.t.} the layer of MLP in the dynamic slot relation completion layer.}
\label{fig:parameter3}
\end{figure}

\begin{figure}[!h]
\centering
\includegraphics[width=0.47\textwidth]{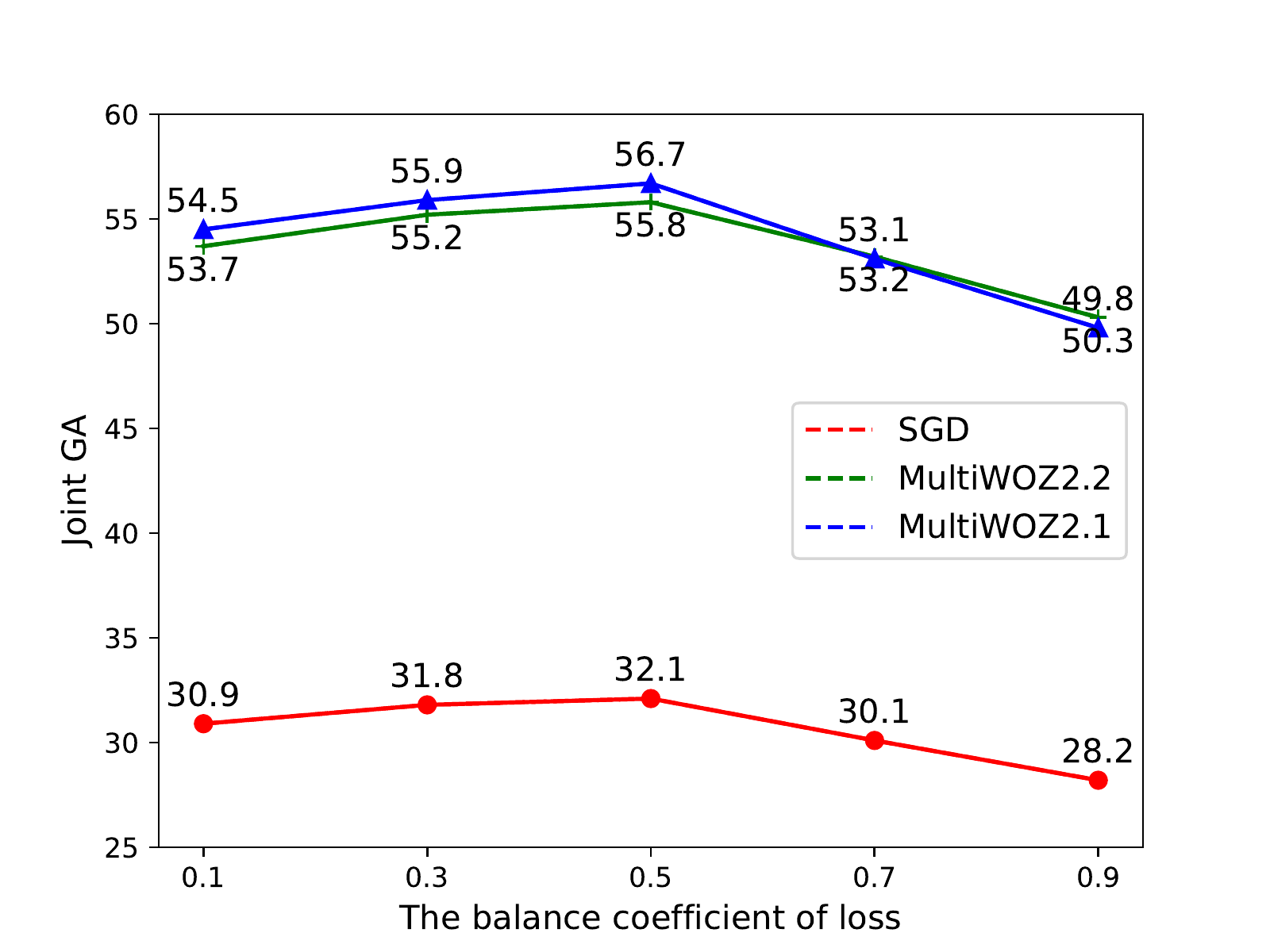}
\caption{Performance comparison \textit{w.r.t.} the balance coefficient in the loss function.}
\label{fig:parameter4}
\end{figure}

We further investigate the impacts of parameter settings on the performance of DSGFNet on SGD, MultiWOZ2.2, and MultiWOZ2.1. We validate the effects of four factors: the layer of propagation on the schema graph, the number of selected dialogue turns used in the schema-dialogue fusion layer, the layer of MLP in the dynamic slot relation completion layer, and the balance coefficient $\lambda$ in the loss function. Figures~\ref{fig:parameter1},~\ref{fig:parameter2},~\ref{fig:parameter3},~\ref{fig:parameter4} show the results of DSGFNet with varying parameters on SGD, MultiWOZ2.2, and MultiWOZ2.1 in terms of joint goal accuracy. We observe that the optimal layer of propagation is not consistent across datasets. It seems that 3 is desired in more datasets. In addition, DSGFNet demonstrates the best performance when leveraging full dialogue history. We conjecture that it is due to that the incomplete dialogue history leads to confusing information. Moreover, 8 layers MLP for relation completion obtains the optimal performance over three datasets. Furthermore, the optimal performance is consistently achieved when the balance coefficient $\lambda$ is around 0.5. 
% We also observe that the performance becomes increasingly worse when $\lambda \rightarrow 0$. This validates the positive benefit of the multi-task learning process in DSGFNet.

\end{document}